\documentclass{ecai}
\usepackage{times}
\usepackage{graphicx}
\usepackage{subfigure}
\usepackage{latexsym}
\usepackage{amsmath,amssymb,amsfonts}
\usepackage[ruled,vlined]{algorithm2e}

\ecaisubmission   

\newcommand{\our}{MIME}

\usepackage{xcolor}
\usepackage{mdframed}

\colorlet{shadecolor}{yellow}

\begin{document}

\title{MIME: Mutual Information Minimisation Exploration}
\author{Haitao Xu \and Brendan McCane \and Lech Szymanski \and Craig Atkinson \institute{Department of Computer Science, University of Otago,
New Zealand, email: \{haitao, brendan, lech, catkinson\}@cs.otago.ac.nz} }

\maketitle
\bibliographystyle{ecai}

\begin{abstract}
  We show that reinforcement learning agents that learn by surprise (surprisal) get stuck at abrupt environmental transition boundaries because these transitions are difficult to learn. We propose a counter-intuitive solution that we call Mutual Information Minimising Exploration (MIME) where an agent learns a latent representation of the environment without trying to predict the future states.  We show that our agent performs significantly better over sharp transition boundaries while matching the performance of surprisal driven agents elsewhere.  In particular, we show state-of-the-art performance on difficult learning games such as Gravitar, Montezuma's Revenge and Doom.
\end{abstract}

\section{Introduction}
%
%
%

Agents trained by reinforcement learning (RL) algorithms perform very well and even exceed human performance in many areas like the game Go \cite{silver2016mastering} and Atari 2600 games \cite{mnih2013playing}. However, the reward function to train an agent is difficult to design and also not scalable as different reward functions are needed for each environment. In addition, some environments only provide extremely sparse rewards.
Random exploration agents, and variants such as $\epsilon$-greedy agents, are not efficient as the agent wastes a lot of time wandering aimlessly over the same states learning nothing, until by chance it hits an extrinsic reward.
It makes more sense to add intrinsic rewards that will encourage efficient exploration; in the absence of extrinsic rewards, the agent is guided to seek new states.

The idea that agents should explore by intrinsic reward like curiosity or surprisal can be traced back to early 1990's. In 1991, Schmidhuber \cite{schmidhuber1991curious} proposed that an agent trained by reinforcement learning can translate mismatches between expectations and reality into curiosity/surprise rewards, also called \textit{surprisal}. The agents are driven to explore surprising aspects of the world, and hence to explore the environment efficiently. This idea has been inherited and carried forward for the next thirty years, especially in recent years. Thanks to increases in computing power, people have verified this idea in large-scale data and scenarios \cite{achiam2017surprise, pathak2017curiosity, burda2018large}.

All of these recent methods follow the same fundamental idea: a reward-maximising neural policy network $\pi$ learns to generate action sequences. A separate neural network called the world model $M$ learns to predict future states ($s_{t+1}$), given past inputs ($s_t$) and actions ($a_t$). In the absence of external reward, the policy network $\pi$ maximises the same value function that the world model $M$ minimises, that is, surprisal rewards. This motivates policy $\pi$ to invent and generate experiments that lead to ``novel" situations where the world model $M$ cannot yet predict well.

However, since the world model $M$ is trained to learn environment transitions, if some of those transitions are discontinuous or abrupt, it is difficult for the agent to predict $s_{t+1}$ and therefore the agent is continuously surprised by the transition. This results in an agent that gets stuck on the transition boundary. The length of time stuck on the boundary depends on the magnitude of change in the transition. The greater the transition change, the longer the time spent at the boundary.

Some current methods tend to avoid getting stuck at transition boundaries.
Houthooft et al. \cite{houthooft2016vime} proposed VIME, which computes Bayesian-surprisal inspired by the idea of maximising information gain. But VIME is difficult to scale up to large-scale environments \cite{achiam2017surprise}. Prediction improvement measures \cite{schmidhuber2006developmental, oudeyer2007intrinsic, lopes2012exploration,achiam2017surprise} compute intrinsic reward by model learning progress and can avoid getting stuck in some situations where surprisal does, for example, when presented with white noise.
Even so, \cite{achiam2017surprise} shows that prediction improvement measures do not explore as well as surprisal in ordinary sparse reward environments.

We propose Mutual Information Minimising Exploration (MIME) in this paper. \our-agents can explore as well as surprisal-agents in sparse reward environments and much better in environments that include abrupt state transitions.

\section{Background}

In RL, at time step $t\geq 0$, the agent is in state $s_t \in \mathcal{S}$, takes an action $a_t \in \mathcal{A}$, receives extrinsic reward $r^e_t$ and transitions to the next state $s_{t+1}\sim P(s_{t+1}|s_t, a_t)$. The objective of RL is to find a policy $\pi$ to maximize the discounted cumulative reward:
\begin{eqnarray}\label{eq:rl}
J = \mathbb{E}_{\pi}[\sum_{t=0}^{\infty}\gamma^tr^e_t],
\end{eqnarray}
where $\pi$ is a mapping (neural network in this paper) from a state to a distribution over actions so that $a_t \sim \pi(\cdot |s_t)$, $\gamma\in [0,1)$ is the discount factor.

When environments only provide extremely sparse rewards, it is hard to use \eqref{eq:rl} to train the policy $\pi$, as almost all $r^e_t=0$. This requires us to add intrinsic rewards $r^i_t$ to encourage the agent to explore efficiently to find the sparse extrinsic rewards. The objective function of RL in \eqref{eq:rl} can be rewritten as:
\begin{eqnarray}\label{eq:rl-int}
J = \mathbb{E}_{\pi}[\sum_{t=0}^{\infty}\gamma^t(r^e_t+\eta * r^i_t)],
\end{eqnarray} where $\eta$ is the trade-off between the intrinsic reward and extrinsic reward. Surprise is a common form of intrinsic reward.

\subsection{Count-based exploration}
The simplest solution for adding intrinsic reward, is to add an intrinsic reward based on a state visit count function:
\begin{equation}\label{eq_statecounting}
r^i_t=\frac{1}{n_t},
\end{equation}
where $n_t$ is the number of times agent has visited state $s_t$ (counting the first visit as $n_t=1$).

This method is obviously only computationally tractable for environments with relatively small number of discrete states.  For continuous state and larger environments something more advanced \cite{ostrovski2017count, bellemare2016unifying} is required.

\subsection{Surprise-driven exploration}
\subsubsection{Surprisal}
Surprise-driven exploration is a class of exploration methods dependent on errors in predicting dynamics \cite{schmidhuber1991possibility, schmidhuber1991curious, schmidhuber2019unsupervised, stadie2015incentivizing, achiam2017surprise, pathak2017curiosity,xu2019vase}. The agent builds a separate neural network called the world model $M$ to predict future state $s_{t+1}$, given current state $s_t$ and current action $a_t$, to approximate the environment transition $P(s_{t+1}|s_t, a_t)$ by $P_M(M(s_t,a_t)|s_t, a_t,\theta)$, where $\theta$ are the parameters of the world model $M$. If the agent finds its prediction $M(s_t,a_t)$ is different from the true future state $s_{t+1}$, it gets surprised.

\begin{figure}
    \centering

    {
        \includegraphics[width=0.85\linewidth]{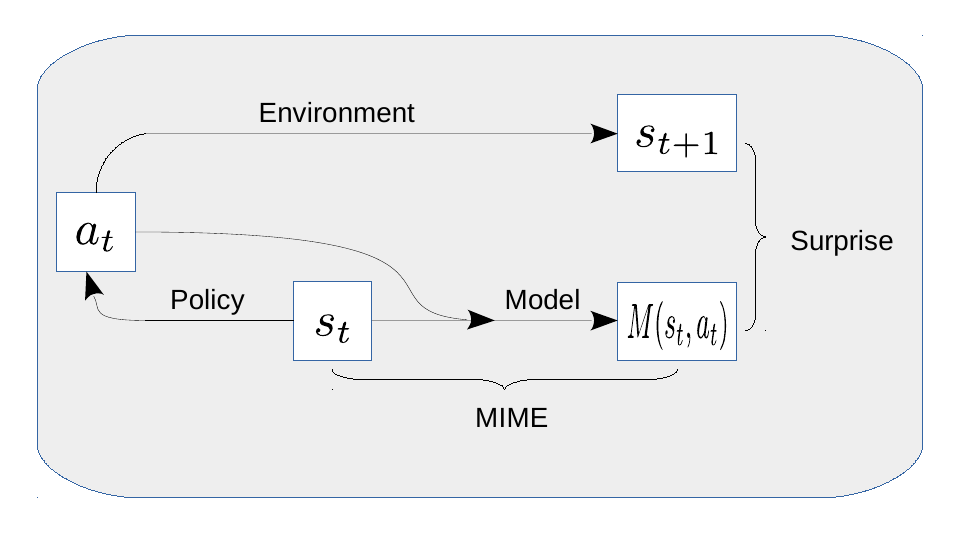}
        \label{fig:tr-rl}
    }
    \caption{surprisal-driven V.S. MIME-driven 
    }
    \label{fig:surprise_reward}
\end{figure}

One simple proposed surprise definition is the so-called surprisal \cite{tribus1961thermostatics}:
\begin{eqnarray}\label{eq:surprisal}
  -\log {P_M(M(s_t,a_t)|s_t, a_t,\theta)}.
\end{eqnarray} In practice, if we suppose $P_M$ is a Gaussian distribution, we can compute the intrinsic reward $r^i_t$ in \eqref{eq:rl-int} with surprisal simply by:
\begin{eqnarray}
r^i_t \sim \|M(s_t,a_t) - s_{t+1}\|^2,
\end{eqnarray}
as shown in Figure \ref{fig:surprise_reward}.

If we train an agent to explore the environment by surprisal, the world model $M$ is trained with the environment transition $P(s_{t+1}|s_t, a_t)$. The world model then learns the mapping function from $(s_t, a_t)$ to $s_{t+1}$ and makes a prediction $M(s_t,a_t)$.  When the environment has an area where the transition $P(s_{t+1}|s_t, a_t)$ from $s_t$ to $s_{t+1}$ is discontinuous, the agent cannot predict $s_{t+1}$ well and gets a big surprise.  The policy network $\pi$ is trained by maximising the surprisal reward, so, at the next iteration, it will generate actions that drive the agent to this area again. Since piecewise smooth functions are hard to learn using continuous activation functions \cite{selmic2002neural}, the agent will be stuck at such transition boundaries for a long time.

VIME agents, as introduced by \cite{houthooft2016vime} can avoid getting stuck at transition boundaries.
VIME agents compute intrinsic rewards based on Bayesian surprise:
\begin{eqnarray}\label{eq:bayesian-surprise}
  r^i_t = D_{KL}[P_M(\theta|s_t,a_t)||P_M(\theta|s_t,a_t,s_{t+1})],
\end{eqnarray}
inspired by maximising mutual information (MI). Since VIME prefers large changes in model parameters from one state to the next, whether it gets stuck at a transition boundary depends heavily on the particular model.
In any case, VIME cannot scale up to large-scale problems that we consider here.

\subsection{Prediction improvement measures}
Another class of intrinsic exploration involves prediction improvement measures \cite{schmidhuber2006developmental, oudeyer2007intrinsic, lopes2012exploration, achiam2017surprise}, which compute the intrinsic reward by:
\begin{eqnarray}
r^i_t &=& \log {P_M(M(s_t,a_t)|s_t, a_t,\theta_t)}\nonumber\\
&&-\log {P_M(M(s_{t-k},a_{t-k})|s_{t-k}, a_{t-k},\theta_{t-k})},
\end{eqnarray} where the intrinsic reward is the $k$-step learning progress at $(s_t, a_t)$. This is somewhat similar to Bayesian surprise and can be easily implemented to large-scale problems, but results in \cite{achiam2017surprise} show that such methods do not perform as well as surprisal.

\section{MIME}
In this section, we present our exploration method named Mutual Information Minimisation Exploration (MIME). Similar to Bayesian surprise and prediction improvement, MIME computes intrinsic reward by mutual information. However,
instead of computing mutual information between past and future steps in the trajectory, we compute mutual information between the input and output of the model $M$ on the current state and current action. In other words, the world model simply tries to learn a representation of the world without prediction, but nevertheless incorporating information from the chosen action. Surprisingly, this works just as well as surprisal in continous transition environments, and much better in environments with abrupt transitions.

The mutual information (MI)-based approach has a long history in unsupervised feature learning and the infomax principle \cite{linsker1988self, bell1995information} for neural networks advocates maximizing the MI between input and output. In our problem, the expected mutual information (information gain) between the input and output of the model $M$ is:
\begin{eqnarray}\label{eq:mi-expected}
  I(\mathcal {S}_{t}\times \mathcal {A}_{t}, M(\mathcal {S}_{t}\times \mathcal {A}_{t})) &\nonumber\\
 =\mathbb{E}_{(s_t,a_t)\sim D}[D_{KL}[P_M(M(s_t,a_t)|s_t,a_t,\theta)||P(M(s_t,a_t))]],&
\end{eqnarray} where $D$ is a dataset of tuples sampled from the environment and used for training.
Since our purpose is to learn the best feature representation to reconstruct $s_t$, we train the model $M$ by maximising \eqref{eq:mi-expected}:

\begin{eqnarray}\label{eq:mi-model}
 \max_{\theta} \mathbb{E}_{(s_t,a_t)\sim D}[D_{KL}[P_M(M(s_t,a_t)|s_t,a_t,\theta)||P(M(s_t,a_t))]].
\end{eqnarray}

Intuitively, one can view the world model as generating a notion of familiarity.
When the agent is in state $s_t$ that it has visited many times before, and chooses action $a_t$ that has been performed in that state before, it ``feels" familiar and comfortable to the agent as the current state and action are well represented. However, the policy network $\pi$ tries to minimise the same mutual information and therefore encourages the agent to either choose actions it has not chosen before, or explore unfamiliar states. In other words, the agent is encouraged to get out of its comfort zone by minimising the mutual information in \eqref{eq:mi-expected}:
\begin{eqnarray}\label{eq:mi-policy}
 \min_{\pi} \mathbb{E}_{(s_t,a_t)\sim \pi}[D_{KL}[P_M(M(s_t,a_t)|s_t,a_t,\theta)||P(M(s_t,a_t))]].
\end{eqnarray}
Minimising equation \eqref{eq:mi-policy} is equivalent to minimising the pair-wise mutual information between the input and output of the model at each step $t$:
\begin{eqnarray}\label{eq:mi-pairwise}
  \log P_M(M(s_t,a_t)|s_t,a_t,\theta)- \log P(M(s_t,a_t)).
\end{eqnarray}

In summary, we train the model $M$ by maximising the mutual information between its input and output to find the best feature representation. In the absence of extrinsic reward, the policy network $\pi$ minimises exactly the same function that the model $M$ is maximising.
To be consistent with the previous reinforcement learning algorithms, the policy is updated by maximizing a reward function. We define the intrinsic reward $r^i_t$ as the negative of \eqref{eq:mi-pairwise}.

\section{Implementation}
Computing the total probability $P(M(s_t,a_t))$ is intractable in practice, so  we only choose the first term from \eqref{eq:mi-pairwise} as an approximate intrinsic reward:
\begin{eqnarray}\label{eq:mi-ri-im}
r^i_t =  - \log P_M(M(s_t,a_t)|s_t,a_t,\theta).
\end{eqnarray}

\noindent Similarly, the objective function in equation \eqref{eq:mi-model} for training the model $M$ can be rewritten as:
\begin{eqnarray}\label{eq:mi-model-im}
    \min_{\theta} -\frac{1}{D}\sum_{(s_t,a_t)\in D}\log P_M(M(s_t,a_t)|s_t,a_t,\theta).
\end{eqnarray}
If we suppose $P_M$ is a Gaussian distribution and that $M$ is autoencoder-like, then $r^i_t$ can be written in a simple way:
\begin{eqnarray}
    r^i_t \sim  \|M(s_t,a_t) - s_{t}\|^2.
\end{eqnarray}

The entire training procedure is summarised in Algorithm \ref{algo:MIME}. Figure \ref{fig:surprise_reward} shows that the structure of surprisal-driven exploration is not changed, just the definition of intrinsic reward.

\begin{algorithm}
\DontPrintSemicolon
\SetAlgoLined
\SetKwInOut{Input}{input}\SetKwInOut{Output}{output}
{Initialise policy neural network $\pi$\;
Initialise world model $M$\;
Reset the environment getting $(s_0,r_0)$\;}
\For{each iteration $n$}
{
    \For{each time step $t$}
    {
    	Get action $a_t \sim \pi(\cdot |s_t)$\;
		Compute intrinsic reward $r^i_t = \|M(s_t,a_t) - s_{t}\|^2$ \;
		Construct cumulative reward $r^e_t + \eta * r^i_t$\;
                Take action $a_t$ getting $(s_{t+1},r^e_{t+1})$\;
	}
	Update $M$ by minimising the sum of $r^i_t$\;
	Update $\pi$ by maximising the sum of $r^e_t + \eta * r^i_t$.
}
\caption{MIME-driven exploration for deep reinforcement learning}\label{algo:MIME}
\end{algorithm}

\section{Experiments}

For illustrative purposes, we begin with two simple experiments and visualise the agent's movements to show its exploration efficiency. Then we implement MIME-driven exploration in three large-scale experiments: Gravitar, Doom, and Montezuma's Revenge. These three games have extremely sparse rewards and are a good test of exploration ability. All large-scale experiments are run three times with different seeds. Table \ref{tb:large_game} shows how we preprocessed the three large-scale environment. We use TRPO \cite{schulman2015trust} in the two simple experiments and PPO \cite{schulman2017proximal} in large-scale games.
\begin{table}
\begin{center}
{\caption{Large-scale games environmental preprocessing.}\label{tb:large_game}}
\begin{tabular}{lc}
\hline
\rule{0pt}{12pt}
Hyperparameter & Setting\\
\hline
Max and skip frames           & 4\\
Grey-scaling                  & True \\
Observation downsampling      & (84, 84)\\
Max episode steps             & 4500\\
Terminal on loss of life      & False\\
\hline
\end{tabular}
\end{center}
\end{table}

In the first two simple experiments, we follow the structure shown in Figure \ref{fig:surprise_reward} and choose rllab \cite{duan2016benchmarking} as the platform to run the code. The model is a simple fully-connected neural network that has one hidden layer of 32 units and the number of units in output layer is equal to the dimension of state.
The hidden layer has rectified linear unit (ReLU) non-linearities. The policy $\pi$ is also a neural network that has one hidden layer of 32 units and tanh nonlinearities. For other hyper-parameter settings please check Table \ref{tb:simple}, where batch size refers to steps collected at each iteration.
\begin{table}
\begin{center}
{\caption{Hyperparameter setting for two simple experiments.}\label{tb:simple}}
\begin{tabular}{lc}
\hline
\rule{0pt}{12pt}
Hyperparameter & Setting\\
\hline
Batch size                          & 5000\\
Max Rollout Length                  & 500 \\
Number of iteration                 & 1000\\
Discount factor                     & 0.99\\

\hline
\end{tabular}
\end{center}
\end{table}

\begin{figure}
    \centering
    \subfigure[MIME-freeze-CNN layers]
    {
        \includegraphics[width=0.85\linewidth]{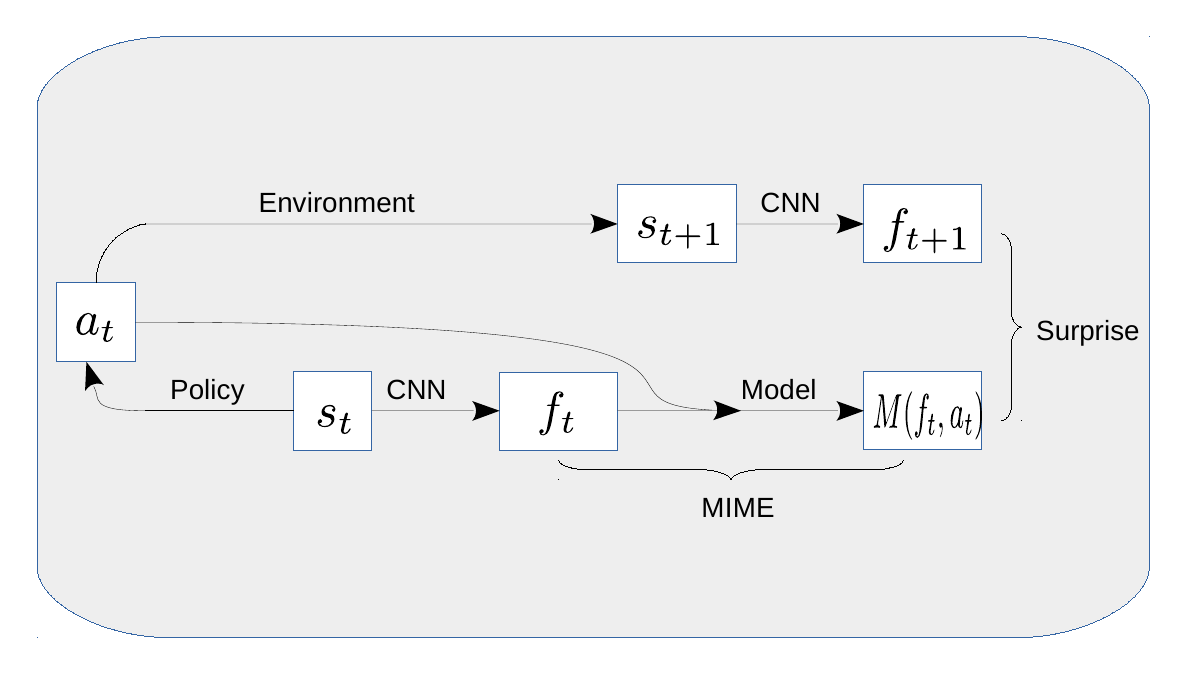}
        \label{fig:surprisal_reward_cnn_1}
    }
    \subfigure[MIME-trainable-CNN layers]
    {
        \includegraphics[width=0.85\linewidth]{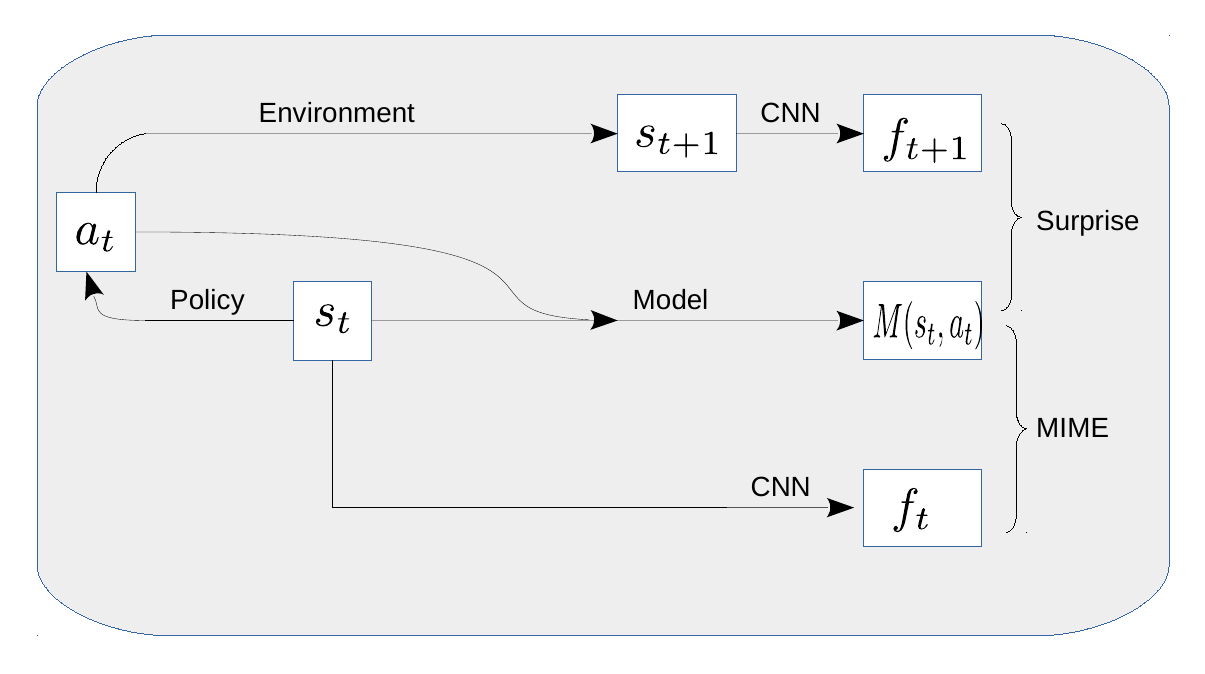}
        \label{fig:surprisal_reward_cnn_2}
    }
    \subfigure[Random network distillation]
    {
        \includegraphics[width=0.85\linewidth]{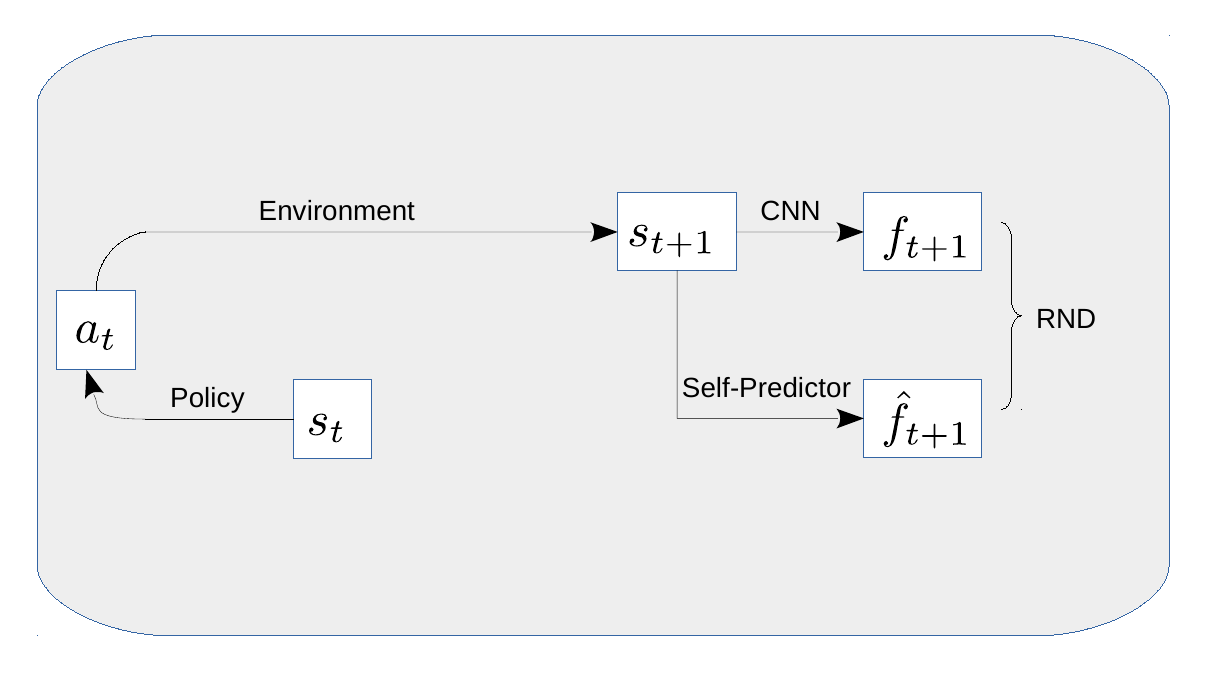}
        \label{fig:RND}
    }
    \caption{Different structures used to compute intrinsic reward.}
    \label{fig:different_structures}
\end{figure}

In the three large-scale games, since the states are pixel-frames, if we also try the structure in Figure \ref{fig:surprise_reward}, we will compute the surprisal or mutual information from raw pixels. However, recent work \cite{pathak2017curiosity, burda2018exploration} shows that if we map the raw pixels to a feature space first and compute the intrinsic reward in this space, we can get a much better result. The mapping can be any feature extractor such as a Variational Autoencoder (VAE) or a Convolution network (CNN) with weights frozen to randomly initialised values.  In this paper, we choose the latter for time efficiency reasons.

We use two different methods to compute MI in a CNN feature space (see Figure \ref{fig:surprisal_reward_cnn_1} and Figure \ref{fig:surprisal_reward_cnn_2}). In Figure \ref{fig:surprisal_reward_cnn_1}, the world model $M$ is trained on the CNN feature space, but in Figure \ref{fig:surprisal_reward_cnn_2}) the world model $M$ is trained from pixels. Both of these two structures use a separate CNN network with frozen layers as a feature extractor so that we can compute intrinsic reward $\|M(f_t,a_t) - f_t\|^2$ or $\|M(s_t,a_t) - f_t\|^2$ in feature space. Another approach called random network distillation (RND) \cite{burda2018exploration} was proposed in 2018 (See Figure \ref{fig:RND}). They chose a self-predictor to predict the features of the next state, the difference between this prediction $\hat f_{t+1}$ and the mapped features $f_{t+1}$ is regarded as the intrinsic reward to train the agent. Here the mapped features $f_{t+1}$ are produced by a CNN with frozen weights. But similar to the surprisal-driven idea, RND also focuses on future states.
We also compare our results with RND in our large-scale experiments.

The feature extractor CNN in Figure \ref{fig:different_structures} has three convolutional layers, which have 32, 64, 64 kernels, 8, 4, 3 kernel size and stride as 4, 2, 1, followed by a fully-connected linear layer with 512 output units. All the parameters in the CNN are fixed (no training). The world model in Figure \ref{fig:surprisal_reward_cnn_1} uses the output of the linear layer as its input and predicts future observations in feature space. It has two hidden non-linear layers and one linear output layer. All these fully-connected layers have 512 output units. The world model in Figure \ref{fig:surprisal_reward_cnn_2} on the other hand, uses the raw pixels as its input. It uses the same convolutional layer structure as the feature extractor CNN we introduced above, but the parameters are trained during learning. The self-predictor in  Figure \ref{fig:RND} has the same structure as the world model in Figure \ref{fig:surprisal_reward_cnn_2}. The difference is that the self-predictor only considers observations as its input and ignores actions. All the policy networks in Figure \ref{fig:different_structures} have the same structure. To compare RND as a baseline, we choose the same other hyper-parameter settings as RND in this paper (see Table \ref{tb:large}). Note that the number of parameters in \ref{fig:surprisal_reward_cnn_1} is much smaller than the other two (3.72 million V.S. 5.14 million).

\begin{table}
\begin{center}
{\caption{Hyperparameter setting for three large-scale games.}\label{tb:large}}
\begin{tabular}{lc}
\hline
\rule{0pt}{12pt}
Hyperparameter & Setting\\
\hline
Rollout length                          & 128\\
Number of iteration                     & 1e8 \\
Number of optimization epochs           & 4\\
Reward trade off $\eta$                 & 0.5\\
Number of parallel environments         & 32 \\
Learning rate                           & 1e-4\\
Discount factor for intrinsic reward    & 0.99\\
Discount factor for extrinsic reward    & 0.999\\
Frames stacked for policy               & 4\\
Frames stacked for model                & 1\\

\hline
\end{tabular}
\end{center}
\end{table}

\subsection{2DPlane environment}\label{sec:2dplane}
\begin{figure}
    \centering
    \subfigure[No intrinsic reward (2,059,459 steps)]
    {
        \includegraphics[width=0.85\linewidth]{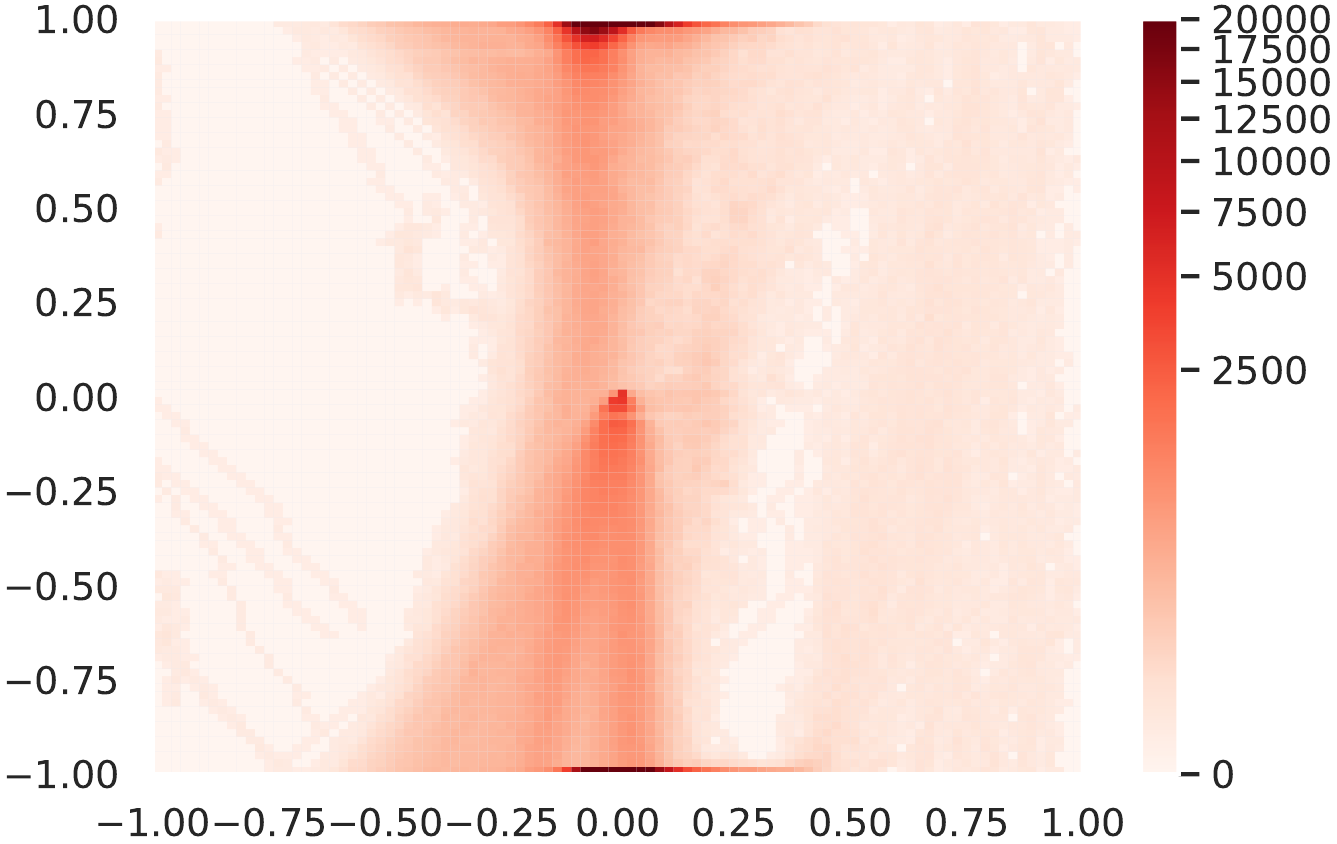}
        \label{fig:exp_strategy_nosup}
    }
    \subfigure[Surprisal (22,919 steps)]
    {
        \includegraphics[width=0.85\linewidth]{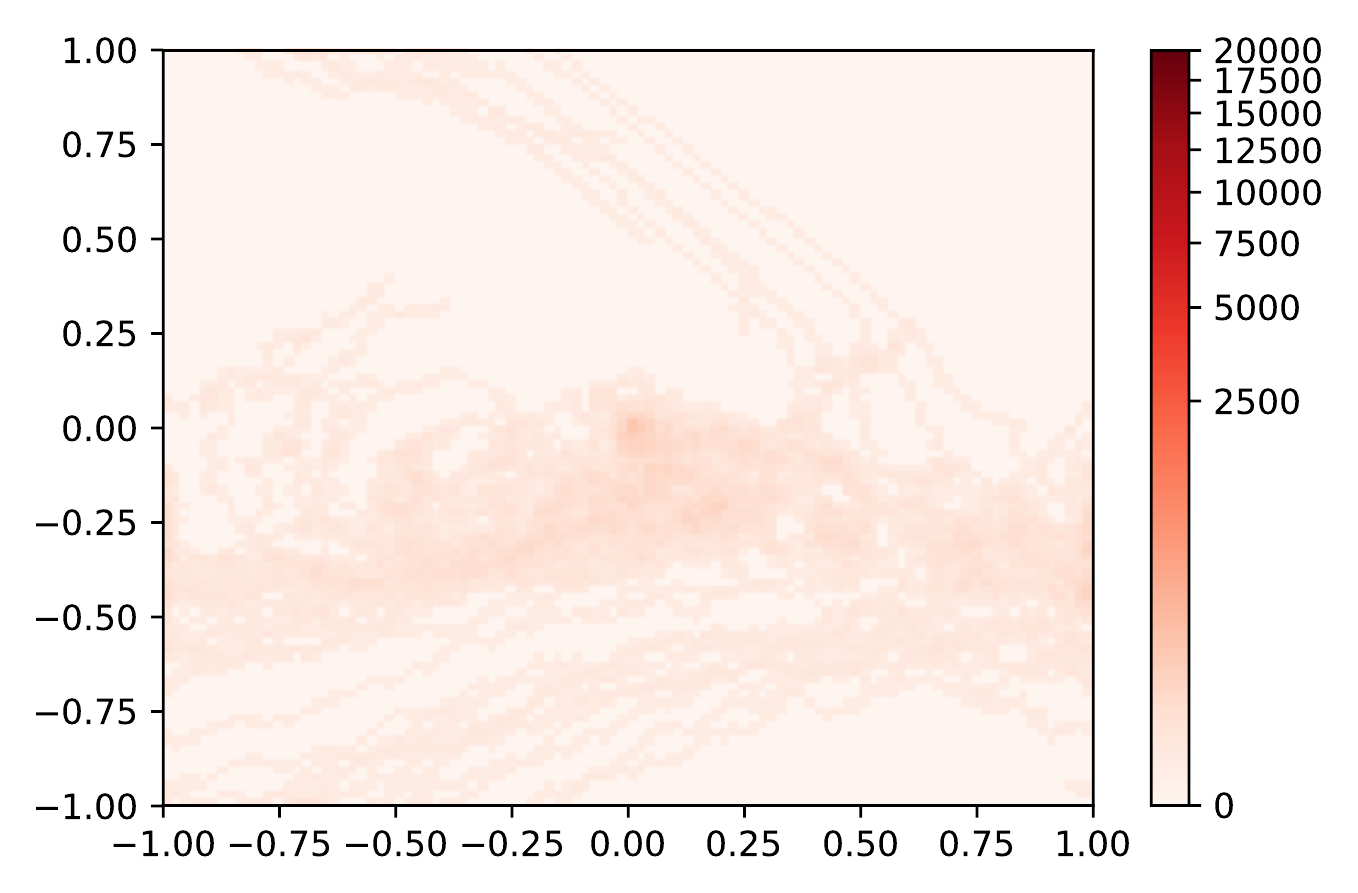}
        \label{fig:exp_strategy_sup}
    }
    \subfigure[MIME (21,150 steps)]
    {
        \includegraphics[width=0.85\linewidth]{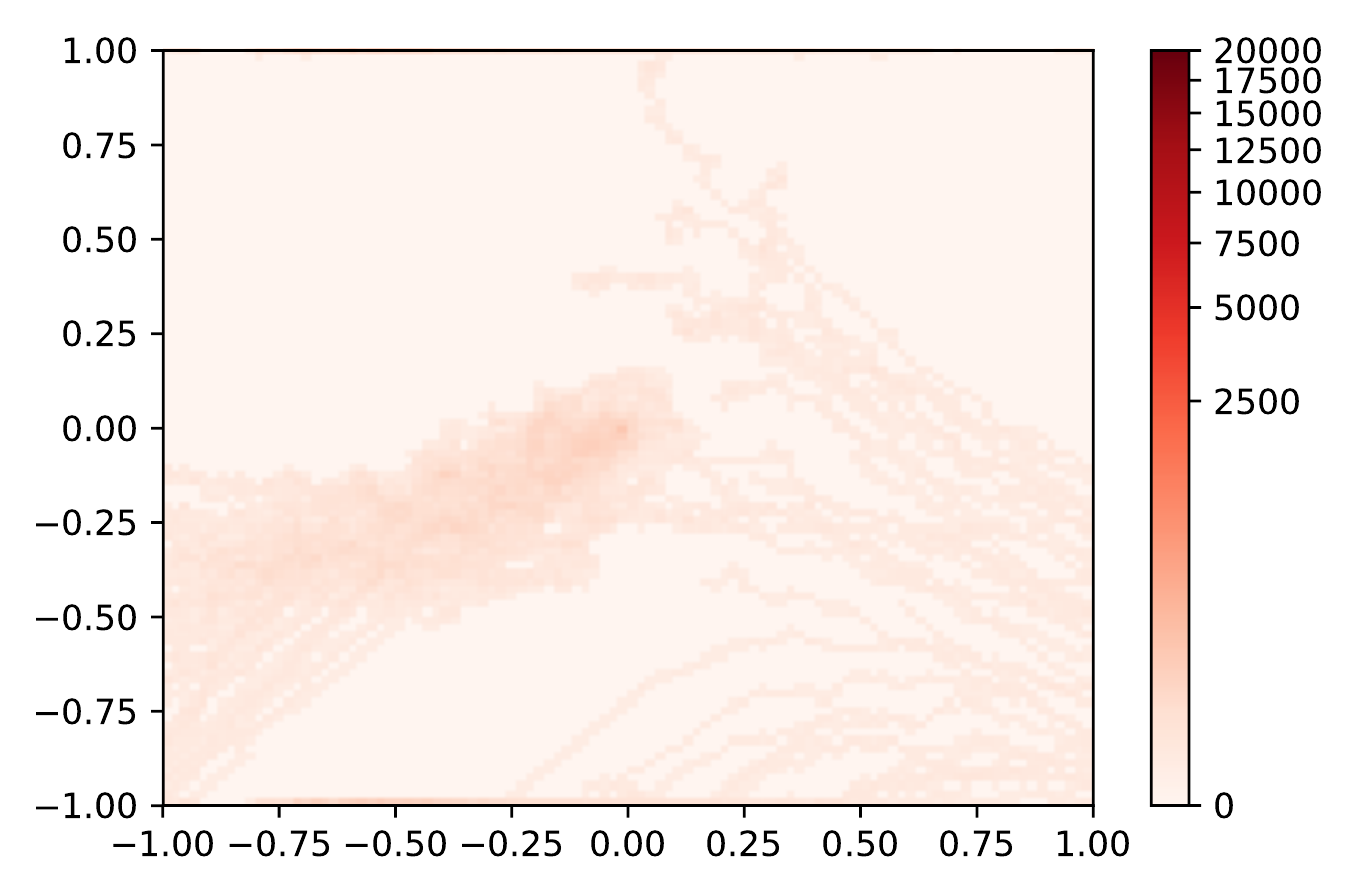}
        \label{fig:exp_strategy_MIME}
    }
    \caption{Exploration efficiency in 2DPlane environment until chancing upon the reward state.}
    \label{fig:exp_strategies}
\end{figure}

This is a simple 2DPlane environment ($\mathcal{S}\subset \mathbb{R}^2, \mathcal{A}\subset \mathbb{R}^2$). We choose this environment to show that MIME-agent has similar exploration ability with surprisal-agent. The observation space is a square on the 2D plane $((x, y) \in \mathbb{R}^2)$, centred on the origin. The action is its velocity $(\dot x, \dot y)$ that satisfies $|\dot x| \leq 0.01, |\dot y| \leq 0.01$. In this environment, the agent starts at origin (0,0) and the only extrinsic reward can be found at location (1,1).  The environment wraps around so that there are no boundaries.

In this experiment, we train one agent and record the observation coordinate $(x,y)$ at each step until it finds the non zero extrinsic reward.  Figure \ref{fig:exp_strategies} shows the heat map of motion tracking for the agent trained without intrinsic reward, with surprisal reward and with MIME reward, respectively. Darker red represents a higher density, which means the agent takes more steps in this area. It is clear that random exploration without intrinsic reward takes a long time (2,059,459 steps) to find the extrinsic reward, whereas surprisal (22,919 steps) and MIME (21,150 steps) do not spend time unnecessarily in random states. We can see that the exploration efficiency is similar between surprisal-agent and MIME-agent, and they are both much more efficient than the random exploration strategy.

\subsection{Passing through a wormhole}\label{sec:wormhole}

\begin{figure}
    \centering
    {
        \includegraphics[width=0.85\linewidth]{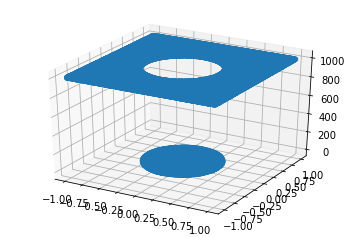}

    }
    \caption{Pass through a wormhole: environment map }\label{fig:wormhole_map}
\end{figure}

\begin{figure}
    \centering
    \subfigure[No intrinsic reward]
    {
        \includegraphics[width=0.85\linewidth]{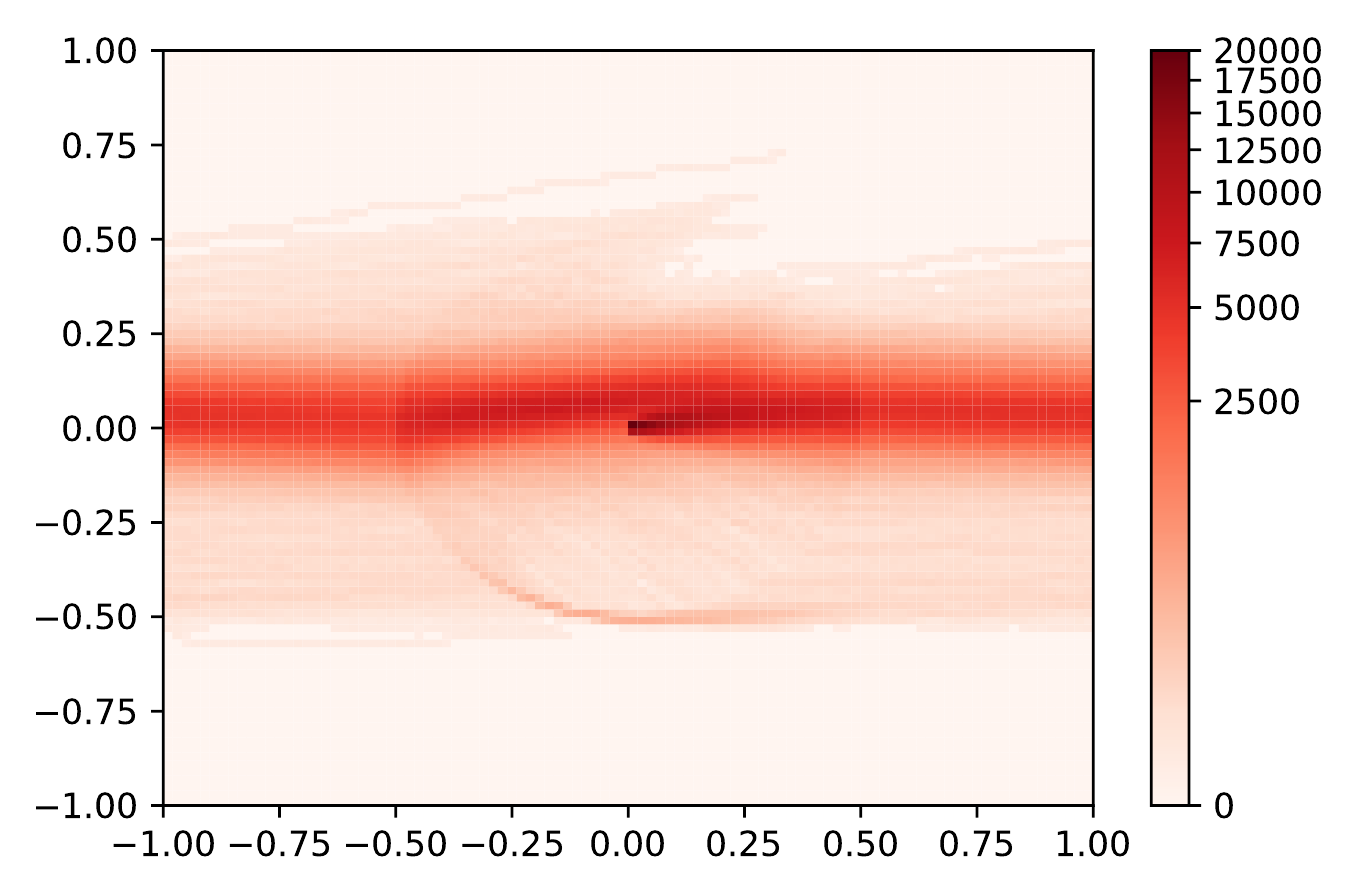}
        \label{fig:no_int_rew_all}
    }
    \subfigure[Surprisal]
    {
        \includegraphics[width=0.85\linewidth]{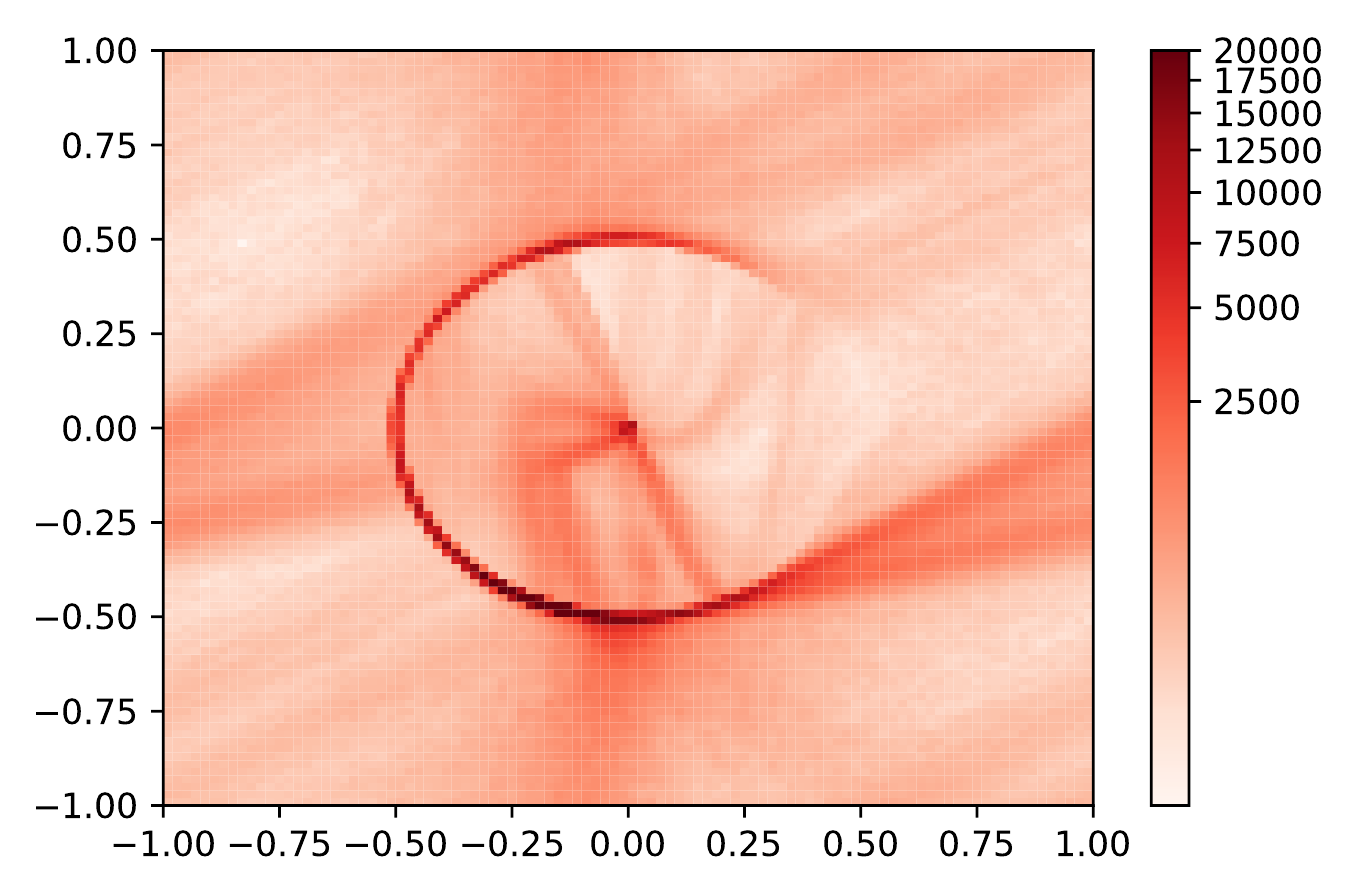}
        \label{fig:wormhole_sup_all}
    }
    \subfigure[MIME]
    {
        \includegraphics[width=0.85\linewidth]{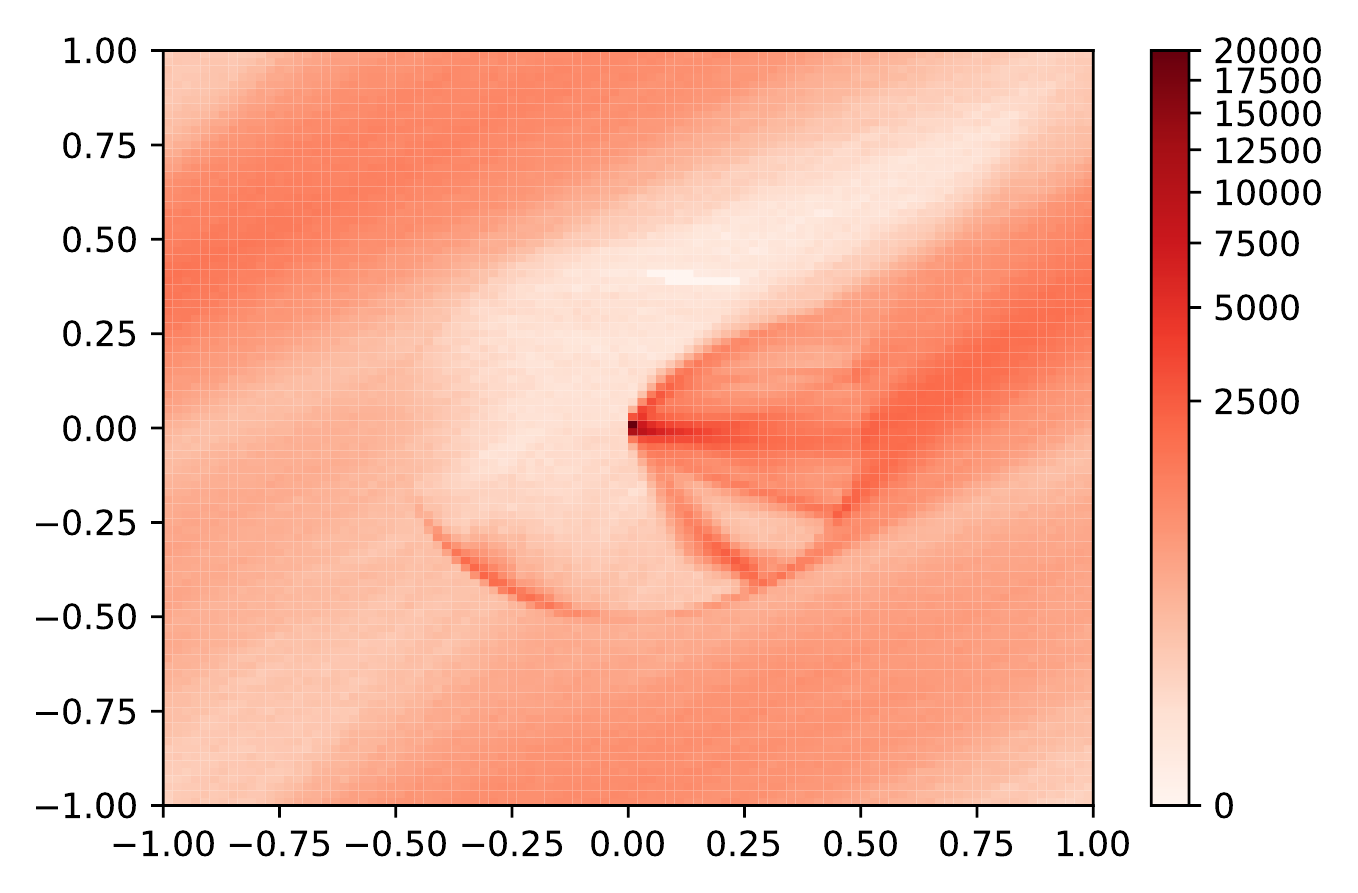}
        \label{fig:wormhole_MIME_all}
    }

    \caption{Pass through a wormhole, 5 million steps. }
    \label{fig:wormhole}
\end{figure}

The wormhole experiment uses a three-dimensional environment with a sharp circular boundary (the wormhole) between an upper rectangular planar environment at $z=1000$, and a lower second circular planar environment centred at the origin with radius $0.5$ (See Figure \ref{fig:wormhole_map}).
The observation space is 3D $((x, y, z) \in \mathbb{R}^3)$. The action is still a two dimensional vector: the velocity $(\dot x, \dot y)$ that satisfies $\sqrt{\dot x^2+\dot y^2} \leq 0.01$.
The agent starts from the origin $(x=0, y=0, z=0)$.
When the agent crosses the boundary, it immediately transitions from one plane to the other.
All the agents explore 5,000,000 steps. Figure \ref{fig:no_int_rew_all} to Figure \ref{fig:wormhole_MIME_all} show the top view of the environment so that we can also visualise the agent's movements. We can see both MIME and surprisal agents are attracted by the boundary, but the time that surprisal agents stay at the boundary is much longer than the MIME agent. As can be seen from the Figure \ref {fig:no_int_rew_all}, the random exploration agent is not affected by the boundary, but the exploration efficiency is very low.

\subsection{Large-scale games}

\begin{figure}
    \centering
    \subfigure[Gravitar]
    {
        \includegraphics[width=0.85\linewidth]{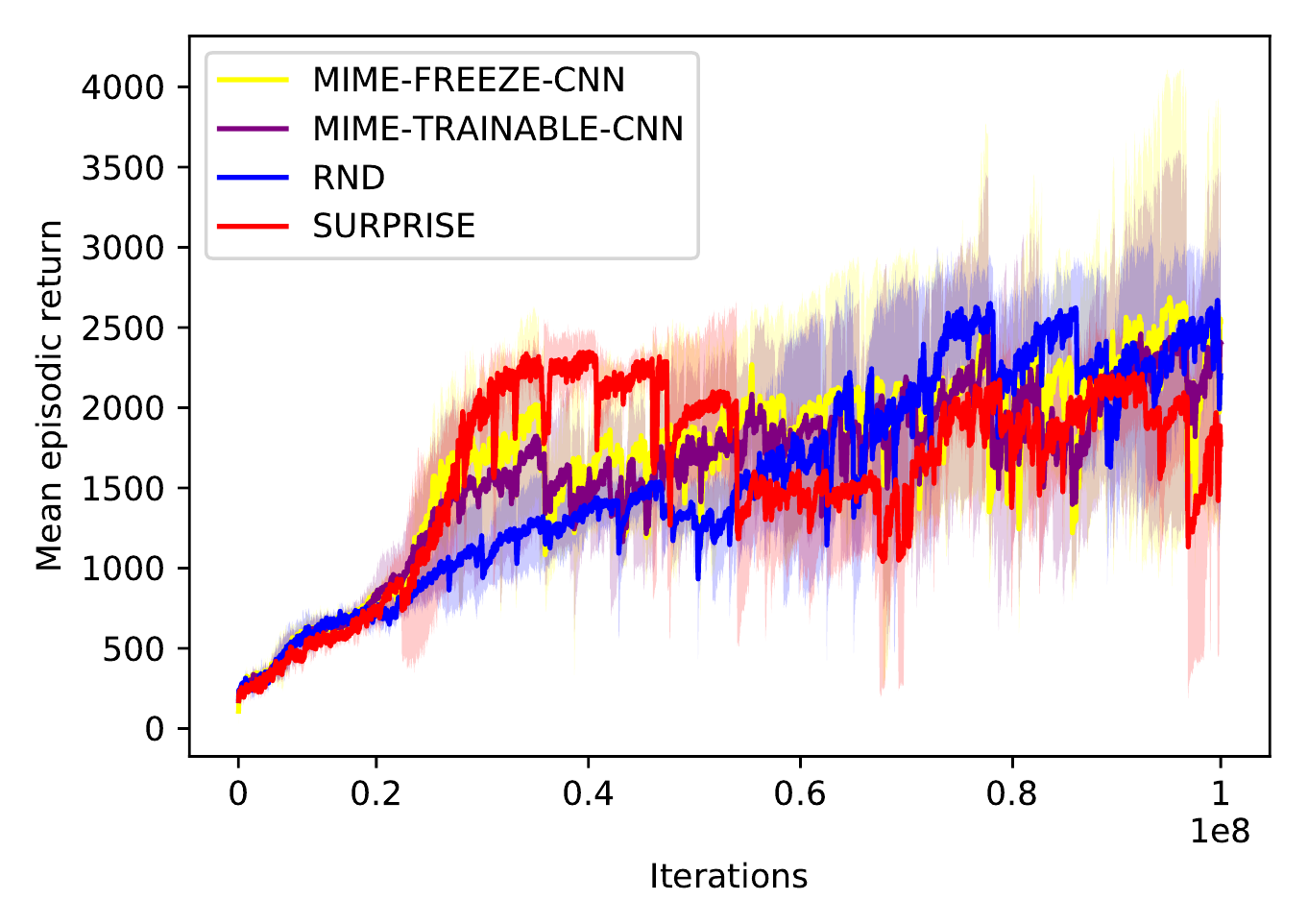}
        \label{fig:gravitar}
    }
    \subfigure[Doom with TV]
    {
        \includegraphics[width=0.85\linewidth]{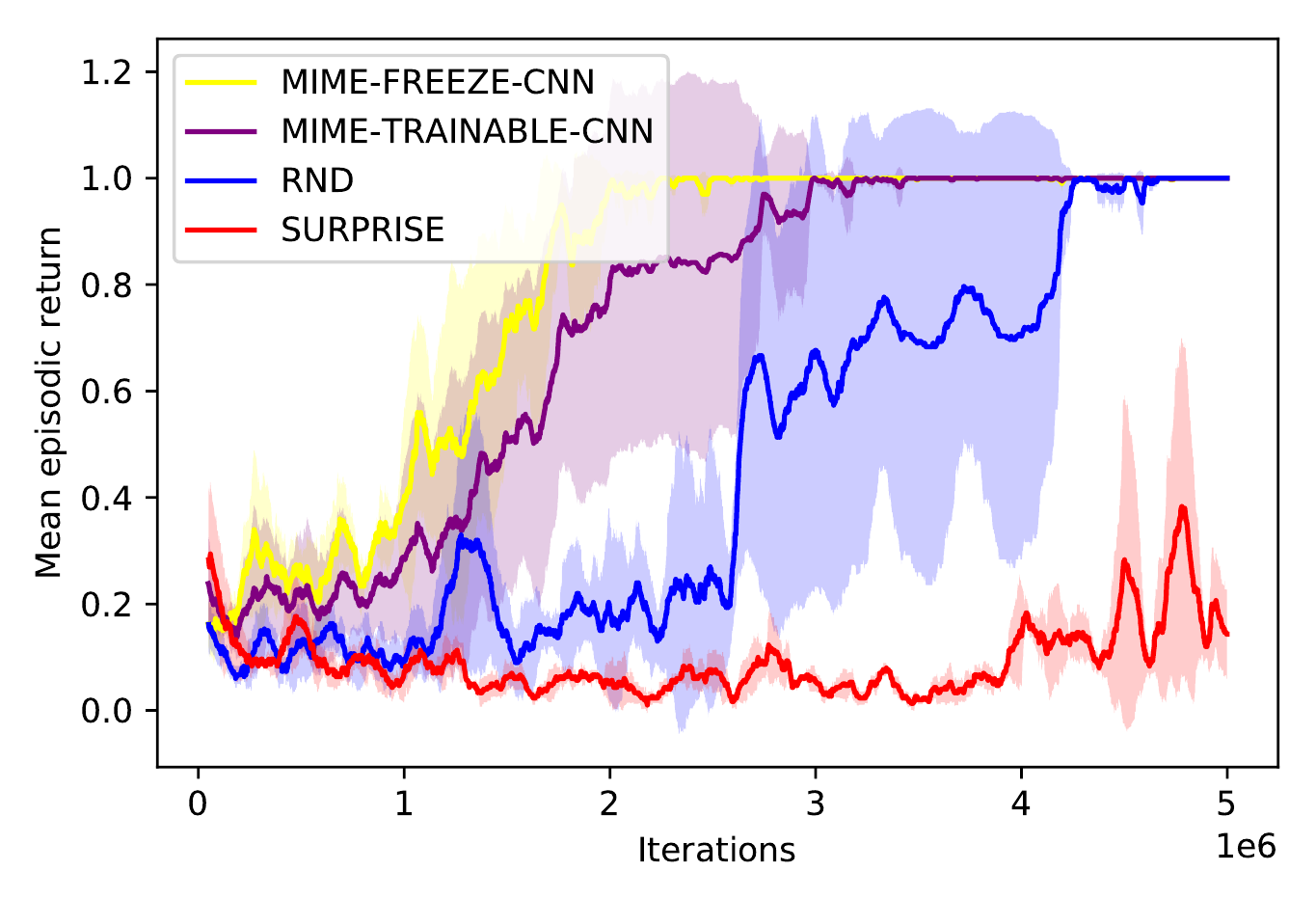}
        \label{fig:doom}
    }
    \subfigure[Montezuma's Revenge]
    {
        \includegraphics[width=0.85\linewidth]{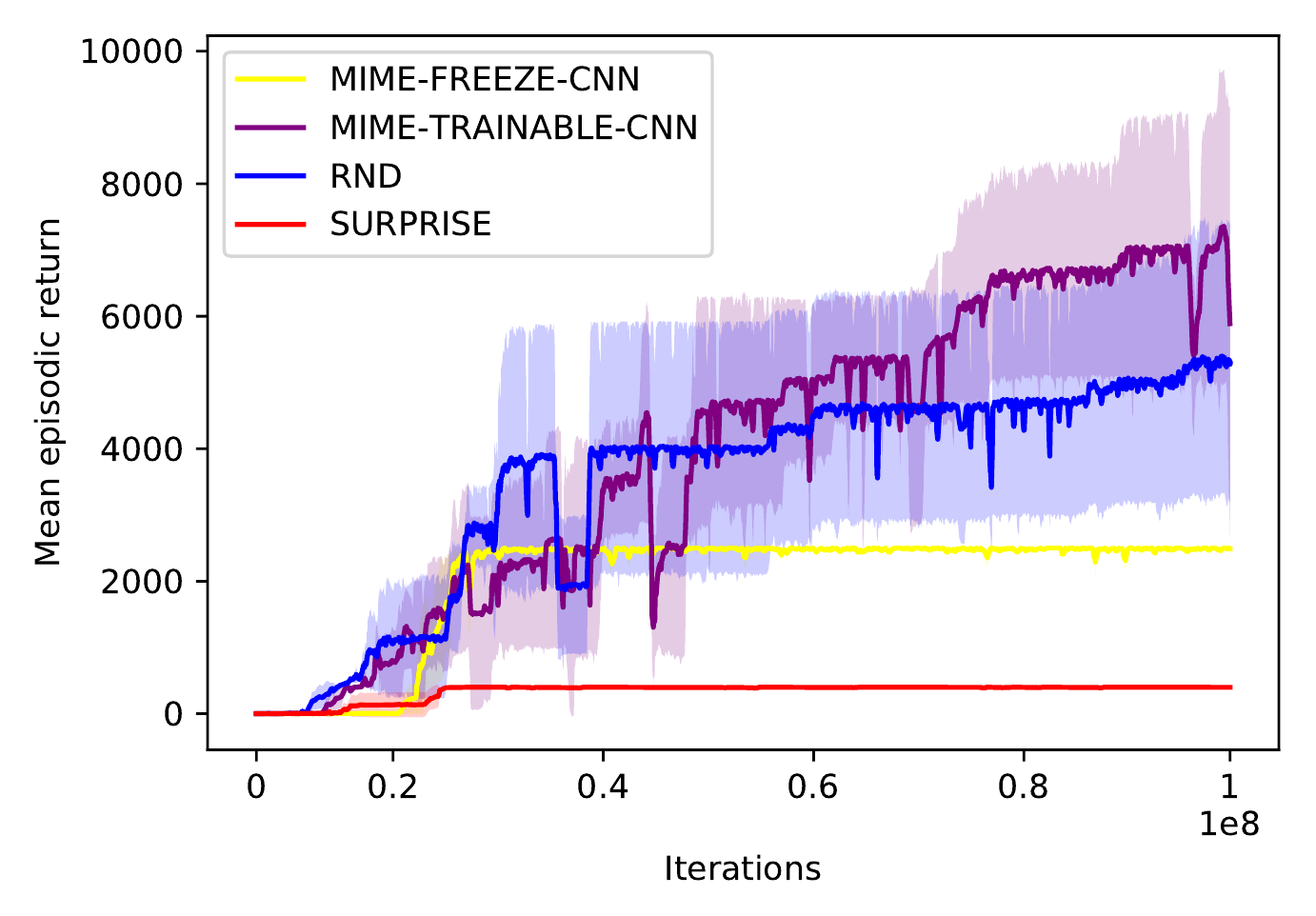}
        \label{fig:montezuma}
    }
    \caption{Mean episodic return of MIME, surprisal, and RND on 3 hard exploration large-scale games. Curves are an average over 3 random seeds, with standard deviation shown in shaded areas. Horizontal axes show numbers of frames.}
    \label{fig:large-scale-game}
\end{figure}

In this subsection, we test MIME and compare it to surprisal and RND in three large-scale games: Gravitar, Doom, and Montezuma's Revenge. For MIME, we implement two different structures as shown in Figure \ref{fig:surprisal_reward_cnn_1} and Figure \ref{fig:surprisal_reward_cnn_2} respectively. All experiments run with 32 parallel environments.

Gravitar is an Atari games which has sparse rewards. There are two modes: the overworld (essentially space); and a sideview landscape when the spaceship enters a planet environment. The agent (spaceship) will be pulled slowly to the star in the overworld, and downward in the side-view levels. We chose this game to show a similar exploration ability between surprisal and MIME. It can be seen from Figure \ref{fig:gravitar} that the agent trained by MIME, surprisal and RND performs similar in this game. The MIME agent that has frozen CNN layers also performs as good as the one with trainable CNN layers.

We choose the scenario named "find my way home" in VizDoom game \cite{wydmuch2018vizdoom} to train the agent to navigate in surroundings and reach his ultimate goal. The map is a series of connected rooms and one corridor with a dead end. Each room has a different colour and texture. There is a green vest in one of the rooms (the same room every time). The agent is born in a randomly chosen room facing a random direction. When the agent explores in this map and finds the vest (the goal), it gets a 1 point reward. We add a TV (always shows changing frames) on the wall in one room and the experiment is run for 5 million frames. We observe that the surprisal agent gets stuck in the TV room, and only the agent born in a room between the TV room and the goal could find the goal, however, both MIME and RND driven agents can escape from the TV room. As with the previous experiment, there is not much difference in the exploration ability between the agents driven from two different structures of MIME (See Figure \ref{fig:doom}).

Montezuma's revenge is an Atari game that is considered very hard. Many RL algorithms \cite{mnih2015human, hessel2018rainbow} that are successful at other Atari games fail at Montezuma's revenge. It has many rooms for the agent to explore. The agent moves from room to room and scores points along the way. This game has similar game mechanics as the wormhole environment we designed in subsection \ref{sec:wormhole}: when the agent moves into a new room, the state abruptly changes because the background of each room is different. Surprisal agents get stuck at the boundary between adjacent rooms because of this transition. In Figure \ref{fig:montezuma} the surprisal agent can only achieve a score of 400 because it is stuck at the boundary between the first and second rooms. A MIME agent with trainable CNN layers performs somewhat better than RND agents. It is interesting to see that MIME agent with frozen CNN layers only scores 2500. The game offers a substantial reward for agents that return to room 2 with the reward from room 3. The frozen CNN MIME agent discovers this reward, and in the process, the policy encourages the agent to go back to room 2 after room 3. We also observe that the RND agent initially gets stuck in the same undesirable loop as the frozen CNN MIME agent. However, the RND agent does manage to escape from this loop in 2 out of 3 of the trails/seeds. We believe that extending training should allow the frozen CNN MIME agent to escape from this loop.

\section{Discussion and Conclusion}
One limitation of our approach is that when we maximise the mutual information, we maximise the KL divergence, which is theoretically without upper bound. For autoencoder-like world models, if the model learns the identity function, the network will be able to reproduce the input regardless of whether it has seen it before. Importantly, this will result in approximately the same intrinsic reward values for every possible observation, regardless of its novelty. In practice, we find that as long as the model contains a layer with significantly fewer units than the input dimensionality, the network does not converge on the identity and thus the intrinsic reward values produced are still useful for preventing the agent from getting stuck in situations described in this paper. We also note that since the world model is trained via mutual information, it need not be autoencoder-like, and could produce output that has a significantly different size to the input.

The main difference between MIME agents and other common RL agents, is that MIME agents do not try to predict the future. Rather, they form a measure of how comfortable they are in a given environment. Whereas surprisal agents explore areas where prediction is poor, MIME agents explore areas where it has a poor world model. As a consequence, surprisal agents tend to seek out hard to learn transition boundaries, whilst MIME agents are encouraged to leave their comfort zone.
This is a simple idea, is easy to implement and most importantly it overcomes the limitations of surprisal getting stuck at transition boundaries.

\ack We gratefully acknowledge the support of NVIDIA Corporation with the donation of the Titan X
GPU used for this research.

\bibliography{ecai}
\end{document}